\documentclass[letterpaper, 10pt, conference]{ieeeconf}

\IEEEoverridecommandlockouts
\usepackage{graphicx}

\newif\ifanonymous
\anonymousfalse   

\ifanonymous
  \usepackage{censor}
\else
  \newcommand{\censor}[1]{#1}
  \newcommand{\blackout}[1]{#1}
  
\fi
\usepackage{cite}
\usepackage{amsmath}
\usepackage{amssymb}
\usepackage{booktabs}
\usepackage[colorlinks=true,linkcolor=black,citecolor=blue,urlcolor=blue]{hyperref}
\makeatletter
\let\NAT@parse\undefined
\makeatother
\usepackage{algorithm}
\usepackage{algorithmic}
\usepackage{subcaption}
\usepackage{siunitx}
\usepackage{multirow}
\usepackage{xcolor}
\usepackage{stfloats}   
\usepackage{tikz}
\usetikzlibrary{calc,arrows.meta,positioning,fit}

\definecolor{goallavender}{HTML}{6A5ACD}
\newlength{\rightH}

\title{\LARGE \bf
  Efficient Autonomous Navigation of a Quadruped Robot in Underground Mines on
  Edge Hardware
}

\author{
  \censor{Yixiang Gao}$^{1}$
  and \censor{Kwame Awuah-Offei}$^{2}$%
\thanks{
  $^{1}$\blackout{Yixiang Gao is with the Department of Mining and Explosive Engineering,
  Missouri University of Science and Technology, Rolla, USA}
{\tt\small \censor{yg5d6@umsystem.edu}}}%
\thanks{
  $^{2}$\blackout{Kwame Awuah-Offei is with the Department of Mining and Explosive Engineering,
  Missouri University of Science and Technology, Rolla, USA}
{\tt\small \censor{kwamea@mst.edu}}}%
}
\begin{document}

\maketitle
\thispagestyle{empty}
\pagestyle{empty}

\ifanonymous\else
  \begin{tikzpicture}[remember picture, overlay]
    \node[anchor=north west, font=\footnotesize\scshape, text=red]
      at ([xshift=1.1cm, yshift=-0.75cm]current page.north west)
      {Preprint, 2026};
  \end{tikzpicture}%
\fi

\begin{abstract}
	Embodied navigation in underground mines faces significant challenges,
	including narrow passages, uneven terrain, near-total darkness, GPS-denied
	conditions, and limited communication infrastructure. While recent
	learning-based approaches rely on GPU-accelerated inference and extensive
	training data, we present a fully autonomous navigation stack for a Boston
	Dynamics Spot quadruped robot that runs entirely on a low-power Intel NUC
	edge computer with no GPU and no network connectivity requirements. The
	system integrates LiDAR-inertial odometry, scan-matching localization 
    against a prior map, terrain segmentation, and visibility-graph global 
    planning with a velocity-regulated local path follower, achieving real-time
	perception-to-action at consistent control rates. After a single mapping 
    pass of the environment, the system handles arbitrary goal locations within 
    the known map without any environment-specific training or learned 
    components. We validate the system through repeated field trials using four 
    target locations of varying traversal difficulty in an experimental 
    underground mine, accumulating over \SI{700}{m} of fully autonomous traverse 
    with a 100\% success rate across all 20 trials (5 repetitions $\times$ 4 
    targets) and an overall Success weighted by Path Length (SPL) of 
    $0.73 \pm 0.09$.
\end{abstract}

\section{INTRODUCTION}
\label{sec:introduction}

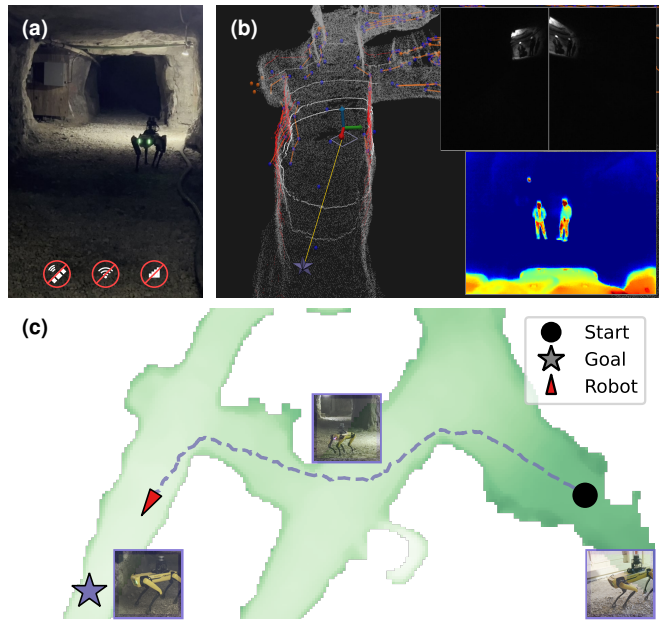
\begin{figure}[t]
	\centering
	\setlength{\rightH}{0.45\columnwidth}
	\begin{tikzpicture}
		\node[anchor=north west, inner sep=0] (photo) at (0,0) {%
			\includegraphics[height=\rightH]{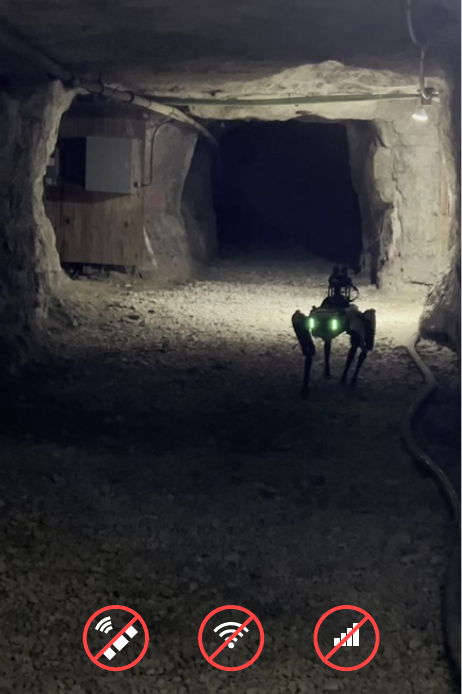}};
		%
		\node[anchor=north west, inner sep=0] (pcd) at
		([xshift=0.02\columnwidth]photo.north east) {%
			\includegraphics[width=0.68\columnwidth, height=\rightH, keepaspectratio=false]{%
				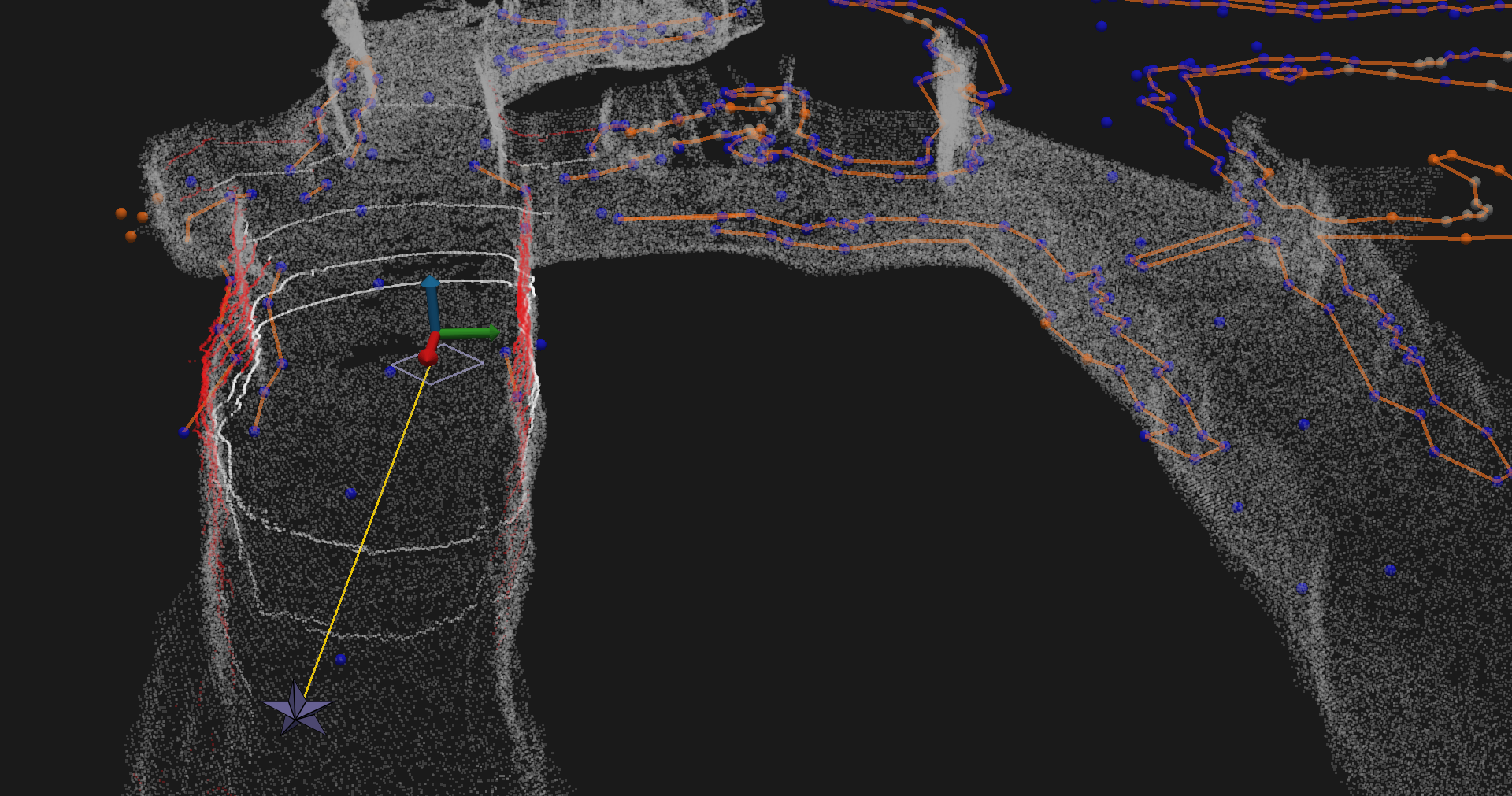}};
		%
		\node[anchor=north east, inner sep=0, draw=gray, line width=0.8pt] (fisheye) at
		([xshift=-1pt, yshift=-1pt]pcd.north east) {%
			\includegraphics[height=0.485\rightH]{%
				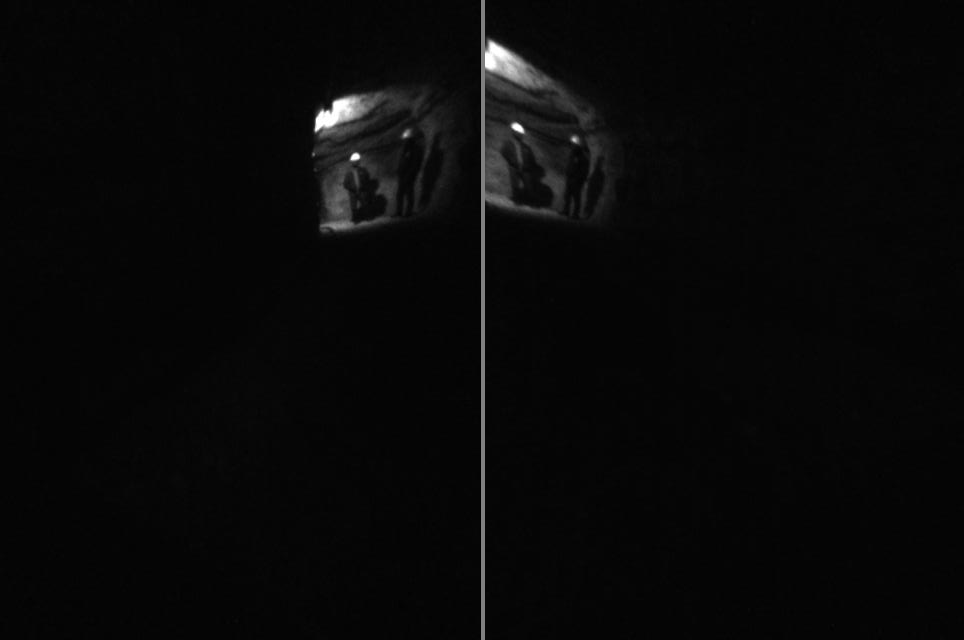}};
		%
		\node[anchor=south east, inner sep=0, draw=gray, line width=0.8pt] (thermal) at
		([xshift=-1pt, yshift=1pt]pcd.south east) {%
			\includegraphics[height=0.485\rightH]{%
				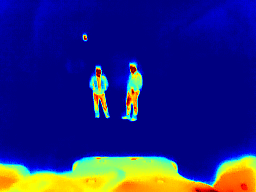}};
		%
		\node[anchor=north west, inner sep=0] (map) at
		([yshift=-0.01\columnwidth]photo.south west) {%
			\includegraphics[width=\columnwidth]{%
				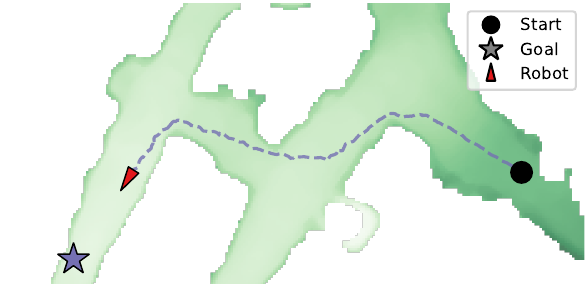}};
		%
		\node[anchor=south east, inner sep=0, draw=goallavender, line width=2pt, opacity=0.8] (entranceview) at
		([xshift=-1pt, yshift=1pt]map.south east) {%
			\includegraphics[height=0.1\columnwidth]{%
				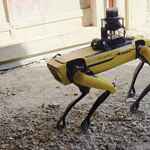}};
		%
		\node[anchor=south west, inner sep=0, draw=goallavender, line width=2pt, opacity=0.8] (robotview) at
		([xshift=115pt, yshift=60pt]map.south west) {%
			\includegraphics[height=0.1\columnwidth]{%
				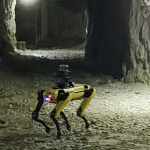}};
		%
		\node[anchor=south west, inner sep=0, draw=goallavender, line width=2pt, opacity=0.8] (goalview) at
		([xshift=40pt, yshift=1pt]map.south west) {%
			\includegraphics[height=0.1\columnwidth]{%
				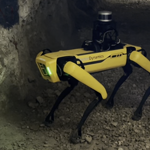}};
		%
		\node[anchor=north west, font=\sffamily\footnotesize\bfseries, text=white]
		at ([xshift=2pt, yshift=-2pt]photo.north west) {(a)};
		\node[anchor=north west, font=\sffamily\footnotesize\bfseries, text=white]
		at ([xshift=2pt, yshift=-2pt]pcd.north west) {(b)};
		\node[anchor=north west, font=\sffamily\footnotesize\bfseries]
		at ([xshift=2pt, yshift=-2pt]map.north west) {(c)};
	\end{tikzpicture}
	\caption{
        (a)~A third-person view of the environment and the robot during an
		autonomous navigation mission in an underground mine with no GPS, WiFi, 
        or cellular signal. (b)~The prior map point cloud overlaid with a 
        real-time LiDAR scan and terrain classification (white = traversable, 
        red = obstacle); insets show the fisheye camera view (near-dark, true 
        lighting) and thermal image. (c)~Localized map with traversed path 
        (dashed) and robot current position; photo insets show the robot at the 
        entrance (right), mid-journey (center), and approaching the goal (left).
    }
	\label{fig:overview}
\end{figure}

Underground mines remain among the most hazardous workplaces worldwide, with
roof falls, toxic gases, and flooding posing persistent risks to human workers.
Autonomous mobile robots can reduce this exposure by performing routine
inspection, mapping, and post-incident reconnaissance without placing personnel
in danger. However, the subterranean mine environment presents a confluence of
challenges rarely encountered together in surface robotics: narrow passages,
uneven terrain with elevation changes, near-total absence of ambient light,
GPS-denied conditions, and limited wireless communication infrastructure.

Recent advances in embodied navigation have been driven predominantly by
learning-based methods. Vision-Language-Action (VLA) models such as
RT-2~\cite{zitkovich2023rt2} and OpenVLA~\cite{kim2024openvla} achieve 
impressive generalization but require billions of parameters and GPU-accelerated 
inference, with latencies of hundreds of milliseconds even on high-end hardware. 
Visual navigation models including ViNT~\cite{shah2023vint} and 
NaVILA~\cite{cheng2025navila} demonstrate learned navigation for mobile and
legged robots, but rely on camera inputs that fail in low-light conditions and
computing resources that exceed edge platforms. These constraints are not
merely inconvenient in underground mines; they are prohibitive: there is no
network to offload computational tasks to, no ambient light for cameras, and no
margin for nondeterministic inference latency on a power budget.

In this paper, we present a fully autonomous navigation system for a Boston
Dynamics Spot quadruped (Fig.~\ref{fig:overview}) that operates entirely on a
low-power Intel NUC edge computer with no GPU and no network connectivity.
The system combines classical, interpretable components---LiDAR-inertial
odometry, scan-matching localization, morphological terrain segmentation,
visibility-graph planning, and velocity-regulated path following---into a
pipeline that achieves deterministic perception-to-action rates on a CPU. After
a single mapping pass of the environment, the system generalizes to arbitrary
goal locations within the known map without any environment-specific training
or domain adaptation. We validate the system through 20 repeated field trials
using four target locations of varying difficulty at the \blackout{Missouri University of
Science \& Technology (Missouri S\&T) Experimental Mine}, demonstrating 100\%
success rate and an SPL of $0.73 \pm 0.09$ over \SI{717}{m} of autonomous
traversal.

The contributions of this paper are:
\begin{itemize}
	\item A fully open-sourced autonomous navigation stack
	    \footnote{Source code and field data: \censor{https://github.com/g1y5x3/spot-edge-nav}}
	    for quadruped robots in underground mines, which runs in real-time on a
	    low-power edge computer without GPU acceleration, network connectivity, 
        or learned components, enabling deployment in infrastructure-denied
	    environments.

	\item Demonstration that the system generalizes to arbitrary goal locations
	    within a known map after a single mapping pass, with no 
        environment-specific training or domain adaptation.

	\item Extensive field validation in a real underground mine across four
	    target locations of varying difficulty (20 trials), with detailed 
        analysis of navigation reliability, path efficiency, and system latency.
\end{itemize}

\section{RELATED WORK}
\label{sec:related_work}

\subsection{Autonomous Navigation in Subterranean Environments}

The DARPA Subterranean Challenge spurred the development of integrated autonomy
systems for underground environments. Because the challenge focused on
\emph{exploration} of unknown spaces with simultaneous artifact detection, the
resulting systems required substantial computational resources:
CERBERUS~\cite{tranzatto2022cerberus} paired onboard CPUs with a Jetson AGX
Xavier GPU for learned traversability and neural-network-based detection; Team
CoSTAR~\cite{agha2022nebula} equipped each ground robot with multiple
high-power processors including GPU modules for semantic scene understanding;
and Team MARBLE~\cite{biggie2023flexible} reported peak power draws of
\SI{425}{W} from a Ryzen Threadripper system with discrete GPUs. Team
Explorer~\cite{cao2022aede} introduced the AEDE planning framework, including
the FAR Planner used in our work. While these systems advanced the state of the
art in subterranean exploration, the operational need in mines is not
exploration but reliable, repeated \emph{point-to-point navigation} on hardware
simple enough for industrial deployment. Our work addresses this need: we
achieve fully autonomous navigation using only a single low-power CPU, with no
GPU, no learned components.

\subsection{Legged Robot Navigation}

Quadruped robots have been increasingly deployed in unstructured environments.
Recent work has explored pairing RGB cameras with learning-based methods to
leverage generalization from pre-trained vision models \cite{fu2022coupling, roth2024viplanner, cheng2025navila}.
However, these methods rely on camera inputs that degrade under poor lighting
and require GPU resources for real-time inference. Miller et al.~\cite{miller2020mine}
demonstrated that LiDAR- and IMU-based systems can effectively navigate in
completely dark areas of underground mines, using autonomous multi-robot
exploration on Boston Dynamics Spot. Our system similarly relies on LiDAR and
IMU sensors, which remain effective regardless of lighting conditions, while
targeting repeated point-to-point navigation in a known map.

\subsection{Localization in GPS-Denied Environments}

Reliable state estimation without GPS is fundamental to subterranean navigation.
LiDAR-inertial odometry methods \cite{xu2022fast_lio2, shan2020liosam} fuse
high-rate IMU data with LiDAR scans to achieve locally consistent, real-time
pose estimation. To correct accumulated drift over long traverses, map-based
localization methods align incoming scans against a prior map; the Normal
Distributions Transform (NDT)~\cite{biber2003ndt} represents the reference map
as a collection of Gaussian distributions and optimizes scan alignment
efficiently. Our system chains these two stages: FAST-LIO2
provides high-rate odometry, and NDT scan matching against a prior point
cloud map provides global drift correction. Because a prior map is already
available from the teleoperated mapping pass, NDT alignment is a natural fit
that yields a globally consistent pose on the edge computer without maintaining
or optimizing a full pose graph.

\subsection{Path Planning for Unstructured Terrain}

Classical planners for unstructured environments range from sampling-based
methods (RRT*, PRM) to graph-based approaches on explicit representations.
The Nav2 framework~\cite{macenski2020marathon2} provides a modular navigation
stack for ROS~2 with plug-in planners and controllers, but its default
configurations and costmap-based planning are primarily validated in structured
indoor settings. FAR Planner~\cite{yang2022far} constructs a dynamic visibility
graph from terrain observations, enabling efficient replanning in environments
with elongated, branching topology such as mine tunnels. For terrain assessment,
elevation mapping~\cite{miki2022elevation} produces high-resolution 2.5D
traversability estimates suited to outdoor terrain with complex geometry.
In mine environments, however, the dominant terrain challenge is distinguishing
flat traversable ground from walls, rubble, and ceiling structures, which is
well addressed by the Progressive Morphological Filter~\cite{zhang2003pmf} that
classifies ground versus obstacle points through iterative morphological
opening. Our system uses PMF's terrain segmentation to distinguish between
ground and obstacles in mine passages, and leverages FAR Planner's visibility
graph for efficient route computation.

\section{SYSTEM OVERVIEW}
\label{sec:system_overview}

\begin{figure*}
	\centering
	\vspace{0.5em}
	\begin{tikzpicture}[
		block/.style={
				rectangle, draw, fill=white,
				minimum height=2em, minimum width=5.5em,
				font=\scriptsize, align=center,
				rounded corners=1.5pt,
			},
		sensor/.style={
				block, fill=blue!8,
				minimum width=3.5em, minimum height=1.6em,
			},
		offline/.style={
				block, dashed, fill=gray!10,
				font=\scriptsize\itshape,
				minimum width=3.5em, minimum height=1.6em,
			},
		input/.style={
				block, fill=orange!12,
				minimum width=3.5em, minimum height=1.6em,
			},
		robot/.style={
				block, fill=green!10,
				minimum width=3.5em, minimum height=1.6em,
			},
		arr/.style={
		-{Stealth[length=4pt]}, semithick,
		},
		topic/.style={
				font=\scriptsize\itshape, text=black!80,
				fill=white, inner sep=1pt,
				align=center,
			},
		freq/.style={
				font=\tiny, text=black!50,
			},
		every node/.append style={inner sep=2.5pt},
		node distance=0.4cm and 0.55cm,
		]

		\node[block] (fastlio) {FAST-LIO2\\[-1pt]{\tiny 10\,Hz}};

		\node[sensor, left=of fastlio, yshift=0.55cm] (lidar) {LiDAR\\[-1pt]{\tiny 10\,Hz}};
		\node[sensor, left=of fastlio, yshift=-0.55cm] (imu) {IMU\\[-1pt]{\tiny 100\,Hz}};

		\def\branchY{0.55cm}   

		\coordinate (split) at ($(fastlio.east)+(0.9cm,0)$);
		\coordinate (merge) at ($(fastlio.east)+(5.0cm,0)$);

		\node[block] (ndt)     at ($(split)!0.5!(merge)+(0,\branchY)$) {NDT\\Localization};
		\node[block] (terrain) at ($(split)!0.5!(merge)+(0,-\branchY)$) {Terrain\\Analysis};

		\path let \p1=(ndt.center), \p2=(terrain.center), \p3=(merge)
		in coordinate (farpos) at (\x3+0.9cm, {0.5*(\y1+\y2)});
		\node[block, anchor=west] (far) at (farpos) {FAR\\Planner\\[-1pt]{\tiny 2.5\,Hz}};

		\node[block, right=0.8cm of far] (pursuit) {Regulated\\Pure Pursuit};
		\node[robot, right=1.2cm of pursuit] (spot) {Spot};

		\node[input, above=0.15cm of ndt, xshift=-1.1cm] (initpose) {Initial Pose};
		\node[offline, above=0.15cm of ndt, xshift=1.1cm] (pcd) {PCD Map};
		\node[offline] (vgh) at (pcd -| far) {Visibility Graph};

		\node[input, below=0.35cm of far] (goal) {Goal Pose};

		\draw[arr] (lidar.east) -- ++(0.275,0) |- ([yshift=3pt]fastlio.west);
		\draw[arr] (imu.east)   -- ++(0.275,0) |- ([yshift=-3pt]fastlio.west);

		\draw[arr] (fastlio.east) -- (split) |- (ndt.west)
		node[topic, pos=0.25, above] {odometry\\{(LIO)}};
		\draw[arr] (fastlio.east) -- (split) |- (terrain.west)
		node[topic, pos=0.25, below] {LiDAR scan\\{(filtered)}};

		\draw[arr] (ndt.east) -- (merge |- ndt.east) |- (far.west)
		node[topic, pos=0.25, above] {odometry\\{(map)}};
		\draw[arr] (terrain.east) -- (merge |- terrain.east) |- (far.west)
		node[topic, pos=0.25, below] {LiDAR scan\\{(segmented)}};

		\draw[arr] (far) -- (pursuit)
		node[topic, midway, above] {path};
		\draw[arr] (pursuit) -- (spot)
		node[topic, midway, above] {velocity};

		\draw[arr] (initpose.east) -| ([xshift=-3pt]ndt.north);
		\draw[arr, dashed] (pcd.west) -| ([xshift=3pt]ndt.north);
		\draw[arr, dashed] (vgh) -- (far);

		\draw[arr] (goal) -- (far);

	\end{tikzpicture}
	\caption{
		System architecture of the autonomous navigation stack. Dashed lines
		indicate offline data loaded at startup. FAST-LIO2 fuses LiDAR and IMU
		inputs to produce odometry and motion-undistorted scans. Two parallel
		branches---NDT localization (drift correction against a prior map) and
		terrain analysis (ground/obstacle segmentation)---feed the FAR Planner,
		which computes a global path tracked by a regulated pure pursuit 
        controller.
	}
	\label{fig:system_architecture}
\end{figure*}
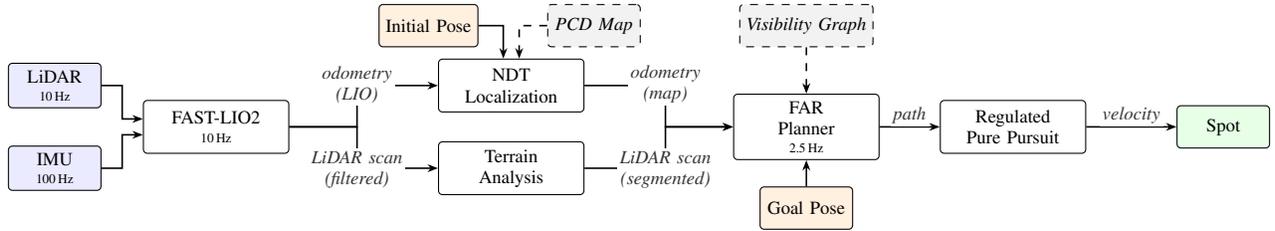

The entire navigation stack is built on ROS~2~\cite{macenski2022ros2} and
consists of four tightly integrated subsystems: (1)~a sensor suite and edge
compute platform, (2)~LiDAR-inertial state estimation with map-based drift
correction, (3)~terrain perception for traversability classification, and
(4)~visibility-graph global planning with regulated local path following.
Fig.~\ref{fig:system_architecture} shows the complete pipeline. Sensor data
flows through FAST-LIO2, which produces both an odometry
estimate and motion-undistorted point clouds. These feed two parallel
branches---NDT localization for global pose correction and terrain analysis for
ground/obstacle segmentation---whose outputs converge at the FAR Planner for
path computation.

\subsection{Hardware Platform}
\label{sec:hardware}

\begin{figure}[thpb]
	\centering
	\begin{tikzpicture}
		\node[anchor=south west, inner sep=0] (main) at (0,0) {%
			\includegraphics[width=0.95\columnwidth, trim=20 0 0 0, clip]{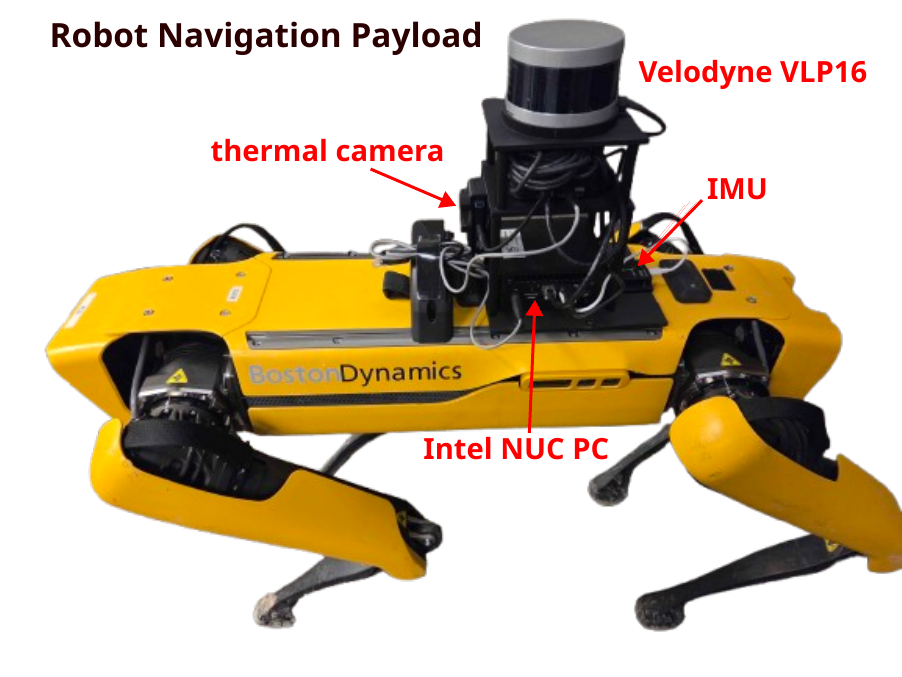}};

		\node[anchor=south east, inner sep=0, draw=gray, line width=1pt] (nuc) at ([xshift=-78pt, yshift=30pt]main.south east) {%
			\includegraphics[width=0.2\columnwidth]{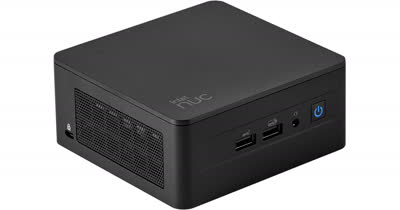}};

		\node[anchor=north west, inner sep=0, draw=gray, line width=1pt] (thermal) at ([xshift=40pt, yshift=-48pt]main.north west) {%
			\includegraphics[width=0.18\columnwidth]{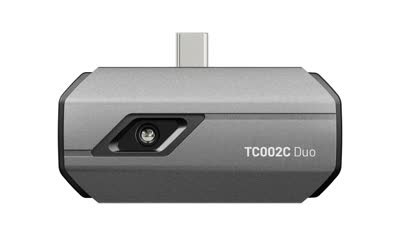}};

		\node[anchor=north east, inner sep=0, draw=gray, line width=1pt] (imu) at ([xshift=-2pt, yshift=-40pt]main.north east) {%
			\includegraphics[width=0.13\columnwidth]{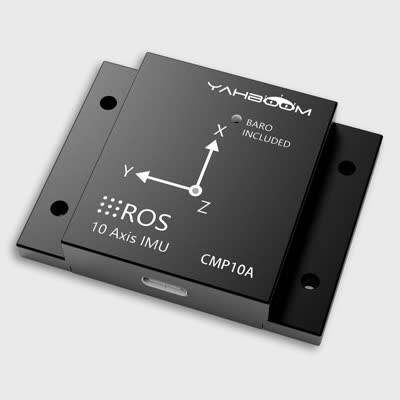}};
	\end{tikzpicture}
	\caption{
		Boston Dynamics Spot with the custom navigation payload. The sensor
		suite includes a Velodyne VLP-16 LiDAR, 
        a Yahboom IMU\protect\footnotemark,
		and a TOPDON TC001 thermal camera. All processing runs onboard on an 
        Intel NUC 13 mini PC.
	}
	\label{fig:hardware}
\end{figure}

The robot platform is a Boston Dynamics Spot quadruped equipped with
a custom sensor payload (Fig.~\ref{fig:hardware}). The sensor suite
consists of:
\begin{itemize}
	\item A \textbf{Velodyne VLP-16} 3D LiDAR mounted at the top of
	      the sensor stack, providing 16-channel point clouds at
	      \SI{10}{Hz} with a \SI{360}{\degree} horizontal field of
	      view (range configured to \SI{0.75}{m}--\SI{30}{m} in
	      software).
	\item A \textbf{Yahboom IMU} providing 6-axis inertial
	      measurements at \SI{100}{Hz} for LiDAR-inertial odometry
	      fusion.
	\item A \textbf{TOPDON TC001 thermal camera} at \SI{30}{Hz},
	      recorded for situational awareness and future use but
	      not part of the autonomous control loop.
\end{itemize}
All computation runs onboard on an Intel NUC 13 (13th Gen i7-1360P, 12 cores /
16 threads, \SI{32}{GB} RAM) configured in performance mode with a sustained
power limit of \SI{40}{W}. Notably, the entire navigation stack runs on CPU
(does not use any discrete GPUs). We do not use Spot's built-in stereo cameras
for navigation due to the low-light mine environment
(see Fig.~\ref{fig:overview},~panel~a and b).

\footnotetext{
    We used an external IMU because accessing the robot's internal 
    high-frequency IMU requires an extra license.
}

\subsection{State Estimation and Localization}
\label{sec:localization}

State estimation follows a two-stage pipeline: high-rate LiDAR-inertial
odometry provides a locally consistent pose, which is then corrected against a
prior map to maintain global accuracy.

\subsubsection{LiDAR-Inertial Odometry}
We employ FAST-LIO2 for real-time LiDAR-inertial odometry. The algorithm fuses
\SI{10}{Hz} LiDAR scans with \SI{100}{Hz} IMU data through an iterated
error-state Kalman filter, performing motion undistortion and point-to-plane
scan matching against an incrementally maintained ikd-Tree map. To meet the
computational budget of the edge computer, raw scans are decimated by a factor
of two and voxel-filtered at \SI{0.5}{m} resolution. The output is 6-DoF
odometry and motion-undistorted registered point clouds at \SI{10}{Hz}.

\subsubsection{NDT Map-Based Localization}
To correct accumulated odometry drift, we apply Normal Distributions Transform
(NDT)~\cite{biber2003ndt} scan matching against a prior point cloud map of the
mine. The prior map is captured in a single teleoperated mapping pass using the
same sensor suite and downsampled to \SI{0.1}{m} voxel resolution at startup.

The localizer maintains a single corrective rigid-body transform
$\mathbf{T}_{\mathrm{map}}^{\mathrm{odom}}$
that relates the FAST-LIO2 odometry frame to the global map frame. At each scan
arrival, the body-frame point cloud is aligned against the NDT voxel grid of
the prior map using the composed prediction
\[
	\mathbf{T}_{\mathrm{map}}^{\mathrm{base}} =
	\mathbf{T}_{\mathrm{map}}^{\mathrm{odom}} \cdot
	\mathbf{T}_{\mathrm{odom}}^{\mathrm{base}}
\]
as the initial guess. If NDT converges (within 30 iterations at \SI{1.0}{m}
voxel resolution), it returns a refined pose
$\hat{\mathbf{T}}_{\mathrm{map}}^{\mathrm{base}}$,
and the correction is updated:
\[
	\mathbf{T}_{\mathrm{map}}^{\mathrm{odom}} \leftarrow
	\hat{\mathbf{T}}_{\mathrm{map}}^{\mathrm{base}} \cdot
	\bigl(\mathbf{T}_{\mathrm{odom}}^{\mathrm{base}}\bigr)^{-1}.
\]
Between NDT updates, the correction is applied to every odometry message,
yielding a globally consistent pose without discontinuities.
\subsection{Terrain Perception}
\label{sec:terrain}

Each registered point cloud from FAST-LIO2 is processed by a terrain
segmentation module that classifies points as traversable ground or obstacle.
First, a ceiling filter discards points above \SI{1.5}{m} in the robot body
frame, removing roof structure and overhead infrastructure that are
prevalent in mine tunnels. The remaining points are then segmented using an
Approximate Progressive Morphological Filter (PMF)~\cite{zhang2003pmf}, which
classifies ground versus obstacle through iterative morphological opening with
parameters tuned for mine geometry. Each point is labeled via the intensity
channel---ground~($0.0$) or obstacle~($1.0$)---and the segmented cloud is
fed to the global planner.

\subsection{Global Planning}
\label{sec:planning}

For global path planning we employ FAR Planner~\cite{yang2022far}, a
visibility-graph-based planner~\cite{cao2022aede}. The planner incrementally
maintains a dynamic visibility graph from the segmented terrain cloud: obstacle
contour polygons are extracted from a bird's-eye occupancy image, convex
vertices are promoted to navigation nodes, and Dijkstra's algorithm computes
shortest paths through observed or unknown space.

In our deployment, a pre-computed visibility graph of the mine
(\texttt{.vgh} file, 846 nodes) is loaded at startup, enabling immediate path
planning without an exploration phase. During traversal, live LiDAR scans
continuously update the graph so that new obstacles (e.g., fallen rock, parked
equipment) not present in the prior map are incorporated and routed around. The
planner runs at \SI{2.5}{Hz} with a sensor range of \SI{30}{m}, robot
radius of \SI{1.0}{m}, and voxel resolution of
\SI{0.05}{m}.

\subsection{Local Path Following}
\label{sec:path_following}

The global path is tracked by a Regulated Pure Pursuit controller~\cite{macenski2023regulated}
with a lookahead distance of \SI{0.5}{m} and maximum linear velocity
of \SI{0.5}{m/s}. The controller computes the curvature of a circular arc to
the lookahead point and regulates speed through curvature scaling, proximity
deceleration near the goal, and a minimum velocity floor to prevent stalling.
The controller runs at the upstream path publication rate (\SI{2.5}{Hz})
with a goal tolerance of \SI{0.3}{m}. Spot's internal locomotion API
interpolates between these velocity commands, so no higher-rate republishing is
needed.

\section{EXPERIMENTAL SETUP}
\label{sec:experiments}

\subsection{Environment}
\label{sec:environment}

We conducted experiments at the \blackout{Missouri S\&T Experimental Mine,
an instructional facility with an underground mine in Rolla, Missouri.} The mine
features narrow passages (approximately \SI{2}{m}--\SI{4}{m} wide),
uneven terrain with mild elevation changes (up to \SI{0.3}{m} over
the full route), multiple T-intersections, and complete absence of
ambient light beyond the first few meters from the entrance. No
GPS, cellular, or WiFi infrastructure is available underground.
The prior point cloud map was captured in a single teleoperated
pass using the same sensor suite and covers approximately
\SI{60}{m} $\times$ \SI{30}{m} of navigable mine passages.

\begin{figure}[thpb]
	\centering
    \vspace{0.5em}
	\includegraphics[width=0.95\columnwidth]{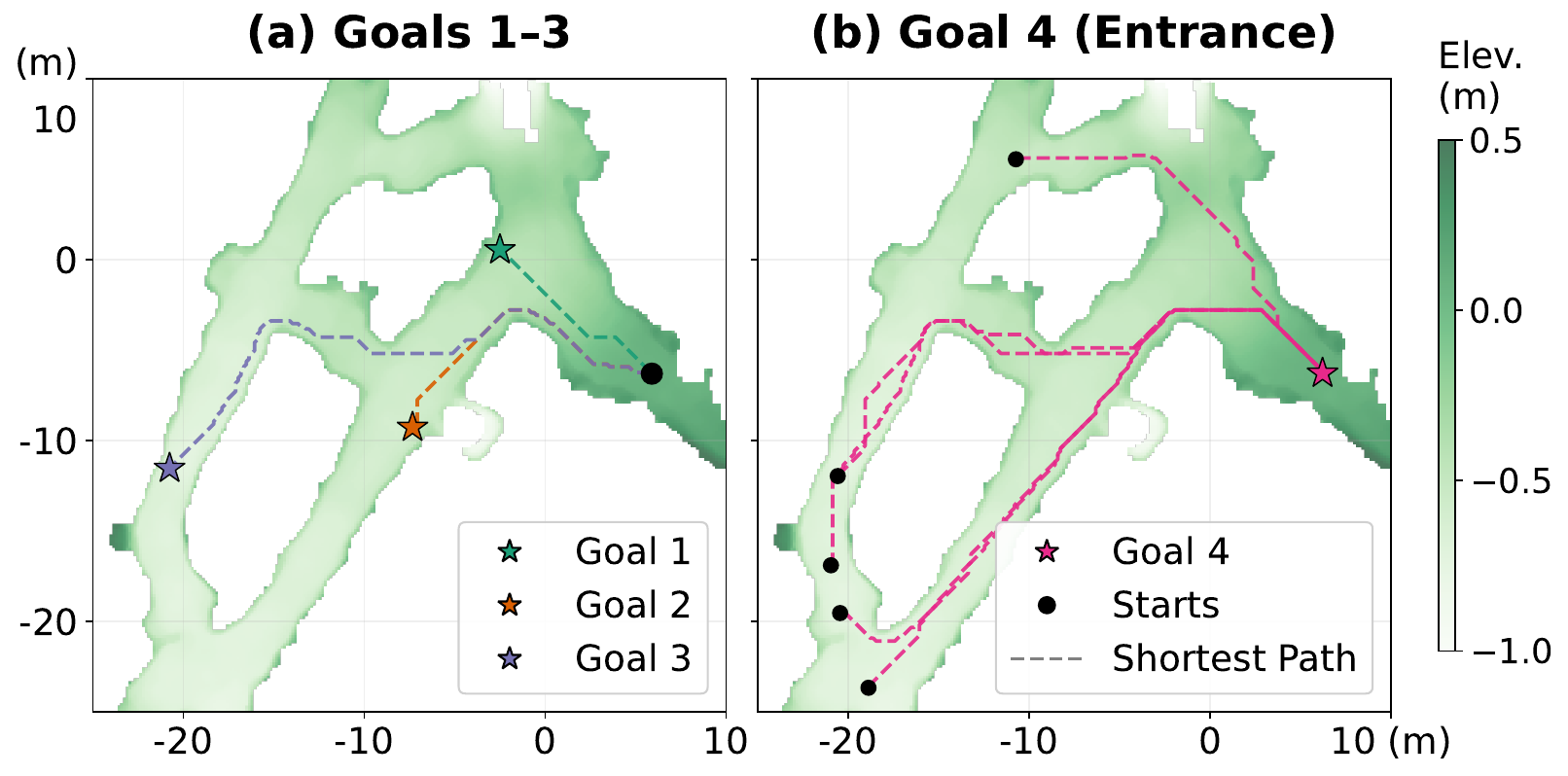}
	\caption{Goal and start locations on the mine elevation
		map (20~trials total).
		(a)~Goals~1--3 share a common start near the entrance.
		(b)~Goal~4 (Entrance): all 5~start locations marked,
		each from a distinct position in the mine.
		Dashed lines show the geodesic shortest paths used to
		compute the reference distance~$\ell$ for SPL evaluation.}
	\label{fig:goals_combined}
\end{figure}

\subsection{Task Description}
\label{sec:task}

The navigation task requires the robot to autonomously traverse
from a known start position near the mine entrance to one of four
predefined target locations of increasing difficulty (Fig.~\ref{fig:goals_combined}):
\begin{itemize}
	\item \textbf{Goal~1 (Easy):} geodesic distance $\ell = \SI{11.2}{m}$,
	      a straight traverse requiring minimal maneuvering.
	\item \textbf{Goal~2 (Intersection):} $\ell = \SI{18.1}{m}$ with
	      \SI{-0.16}{m} elevation change, through a mine intersection
	      requiring turn decisions.
	\item \textbf{Goal~3 (Deep):} $\ell = \SI{34.9}{m}$ with
	      \SI{-0.32}{m} elevation change, deep into the mine.
	\item \textbf{Goal~4 (Arbitrary Start $\to$ Entrance):}
	      $\ell = \SI{35.1}{m}$ (average), starting from five distinct
	      positions throughout the mine and navigating to the entrance.
	      This tests generalization beyond a fixed start.
\end{itemize}
We repeated each experiment 5 times for a total of 20 trials.
A trial is successful if the robot reaches within \SI{1.0}{m}
of the target without human intervention. Human observers, stationed at each goal location, verified success.

\section{RESULTS}
\label{sec:results}
\begin{figure*}[t]
	\centering
    \vspace{0.5em}
	\begin{tikzpicture}[
			imagebox/.style={
					anchor=north west, inner sep=0, outer sep=0
				},
			labelbox/.style={
					anchor=north west, align=left,
					fill=black!80, text=white, font=\sffamily\footnotesize, inner sep=2pt, opacity=0.8, text opacity=1
				},
			labelboxbl/.style={
					anchor=south west, align=left,
					fill=black!80, text=white, font=\sffamily\footnotesize, inner sep=2pt, opacity=0.8, text opacity=1
				},
		]
		\def\W{0.165\textwidth}        
		\def\Hphoto{0.0928\textwidth}  
		\def\Hmap{0.099\textwidth}     

		\def\Cone{0.00\textwidth}
		\def\Ctwo{0.165\textwidth}
		\def\Cthree{0.33\textwidth}
		\def\Cfour{0.505\textwidth}
		\def\Cfive{0.670\textwidth}
		\def\Csix{0.835\textwidth}

		\def\PAone{0.00\textwidth}
		\def\PAtwo{-0.09\textwidth}

		\node[imagebox] (a1c1) at (\Cone, \PAone)
		{\includegraphics[width=\W, height=\Hphoto]{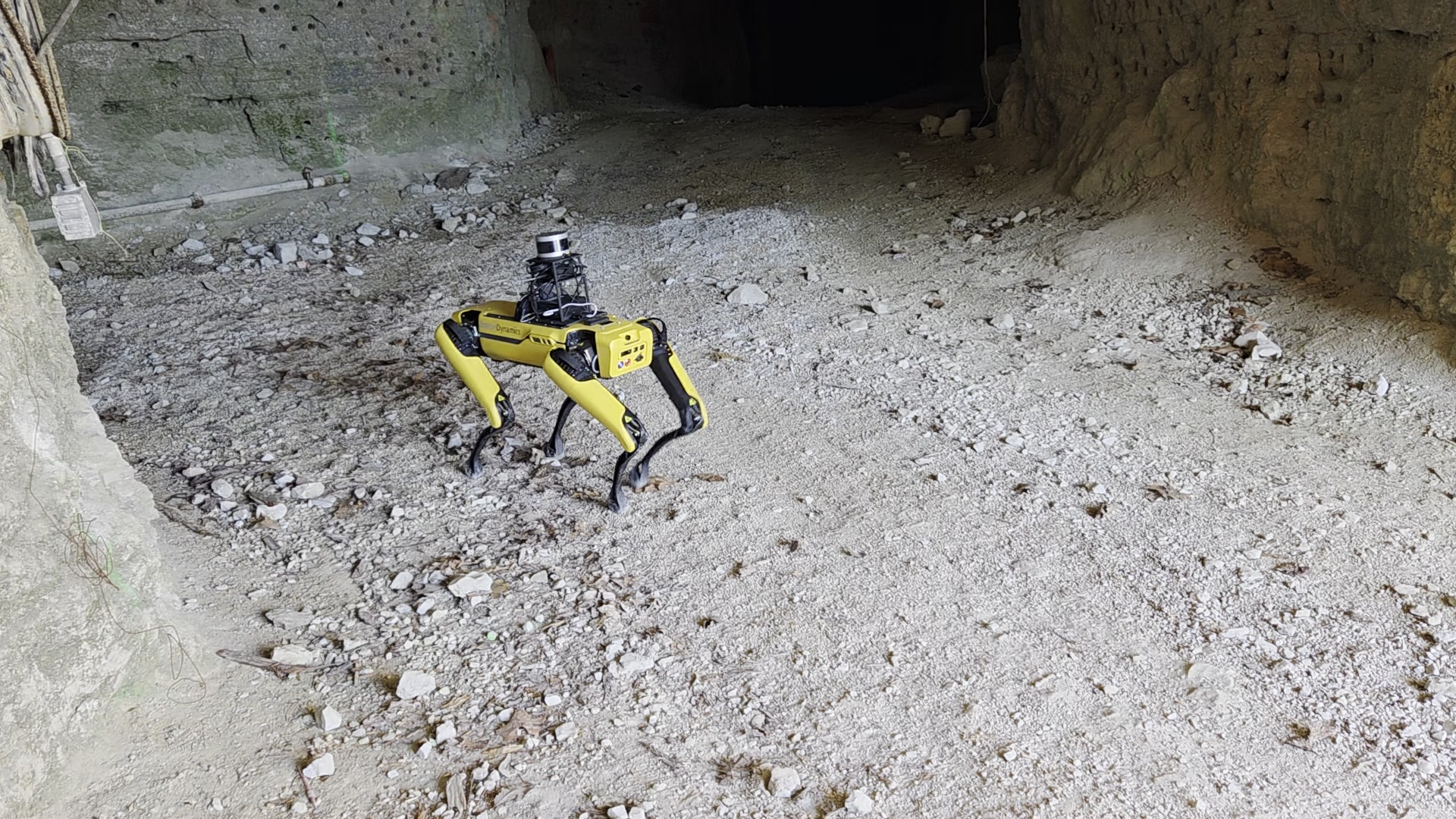}};
		\node[labelbox] at (a1c1.north west) {(a) $t_1$: Entrance};

		\node[imagebox] (a2c1) at (\Cone, \PAtwo)
		{\includegraphics[width=\W, height=\Hmap]{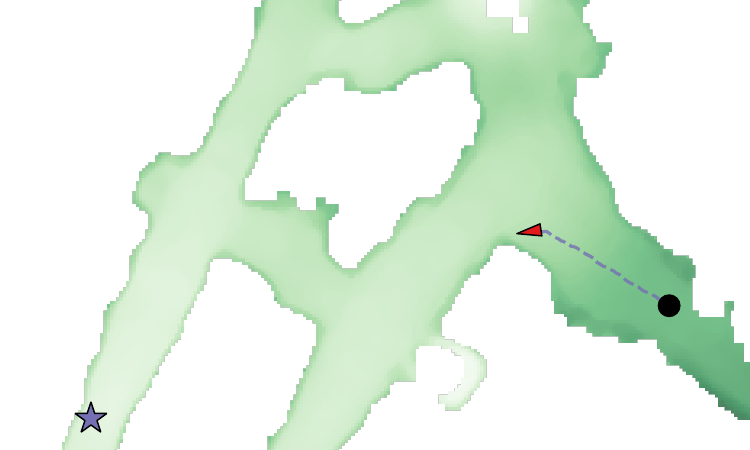}};
		\node[labelbox] at (a2c1.north west) {Elapsed: 19.19\,s\\Dist: 8.35\,m};

		\node[imagebox] (a1c2) at (\Ctwo, \PAone)
		{\includegraphics[width=\W, height=\Hphoto]{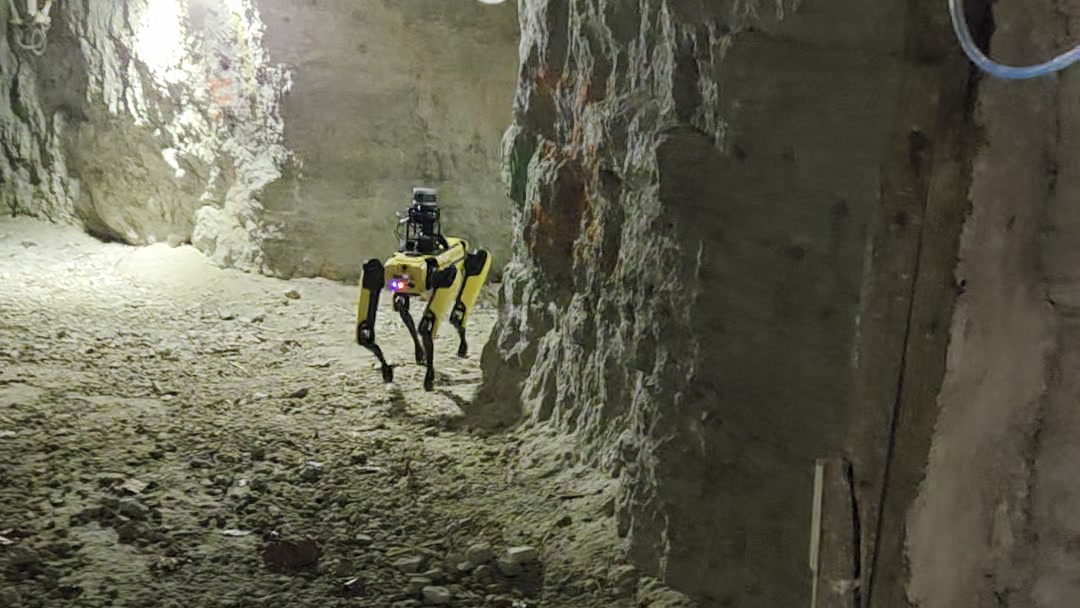}};
		\node[labelbox] at (a1c2.north west) {(b) $t_2$: Mid-Journey};

		\node[imagebox] (a2c2) at (\Ctwo, \PAtwo)
		{\includegraphics[width=\W, height=\Hmap]{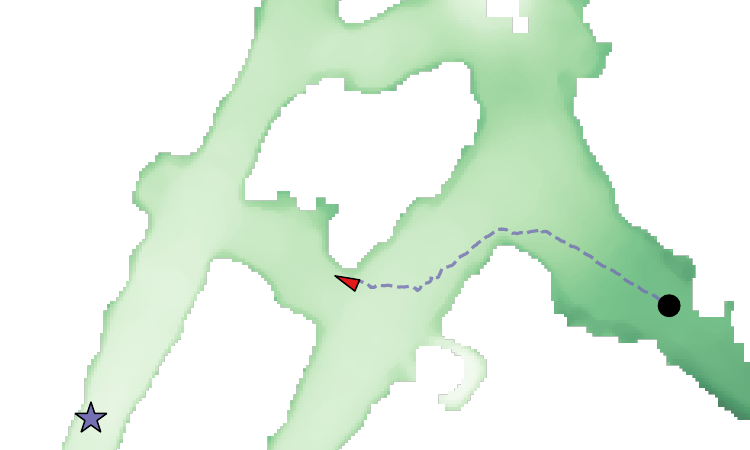}};
		\node[labelbox] at (a2c2.north west) {Elapsed: 46.19\,s\\Dist: 18.88\,m};

		\node[imagebox] (a1c3) at (\Cthree, \PAone)
		{\includegraphics[width=\W, height=\Hphoto]{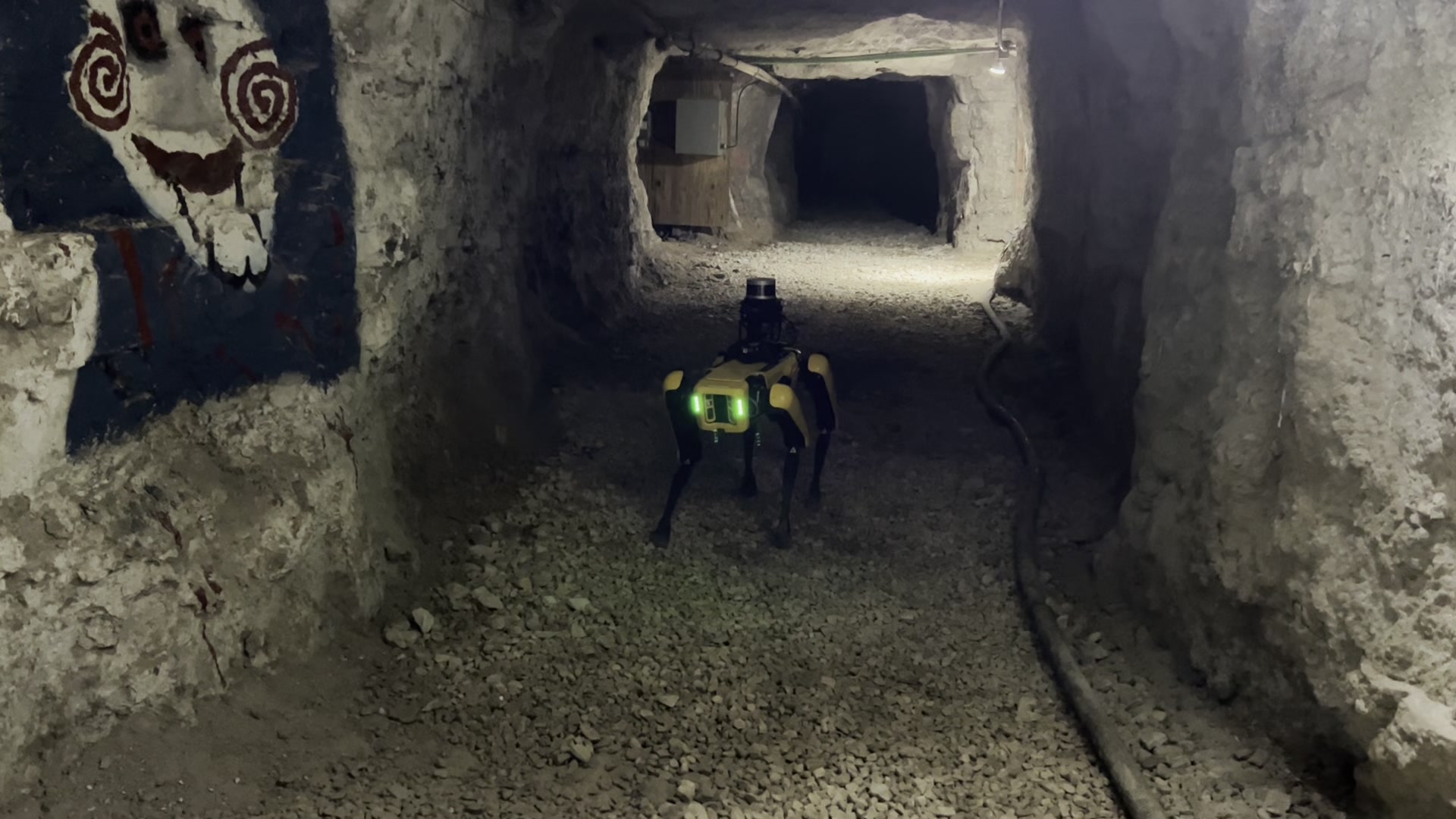}};
		\node[labelbox] at (a1c3.north west) {(c) $t_3$: Goal Arrival};

		\node[imagebox] (a2c3) at (\Cthree, \PAtwo)
		{\includegraphics[width=\W, height=\Hmap]{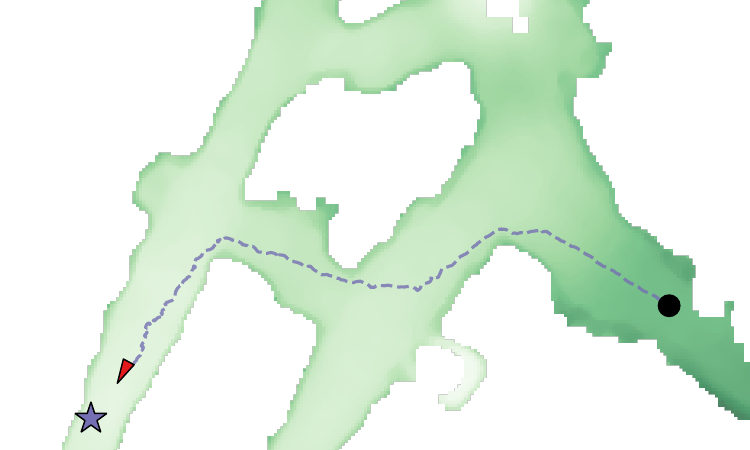}};
		\node[labelbox] at (a2c3.north west) {Elapsed: 100.19\,s\\Dist: 36.65\,m};


		\node[imagebox] (b1c1) at (\Cfour, \PAone)
		{\includegraphics[width=\W, height=\Hphoto]{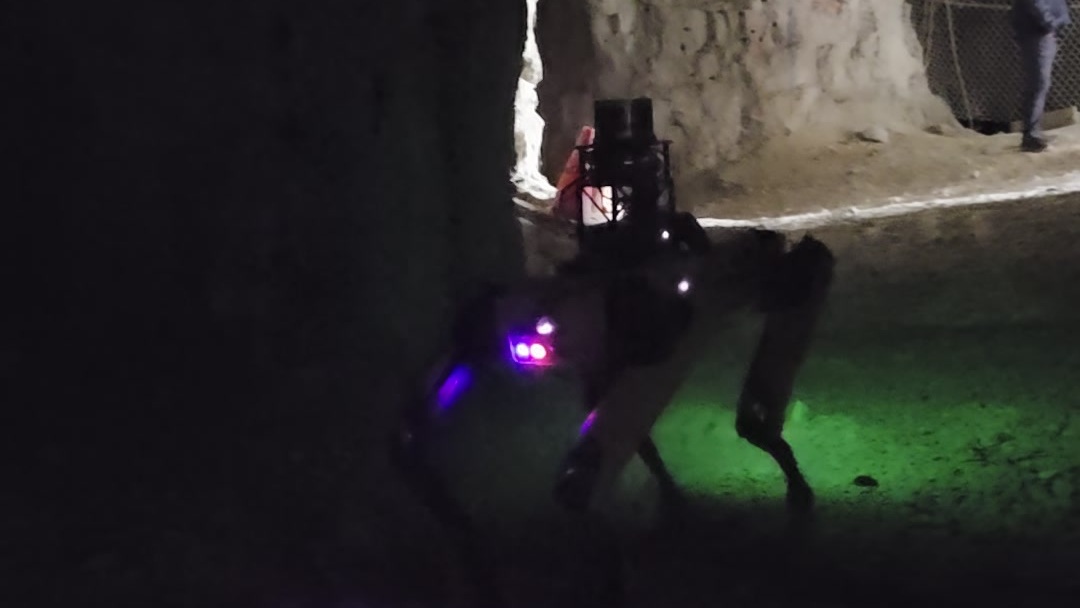}};
		\node[labelboxbl] at (b1c1.south west) {(d) $t_1$: Deep Start};

		\node[imagebox] (b2c1) at (\Cfour, \PAtwo)
		{\includegraphics[width=\W, height=\Hmap]{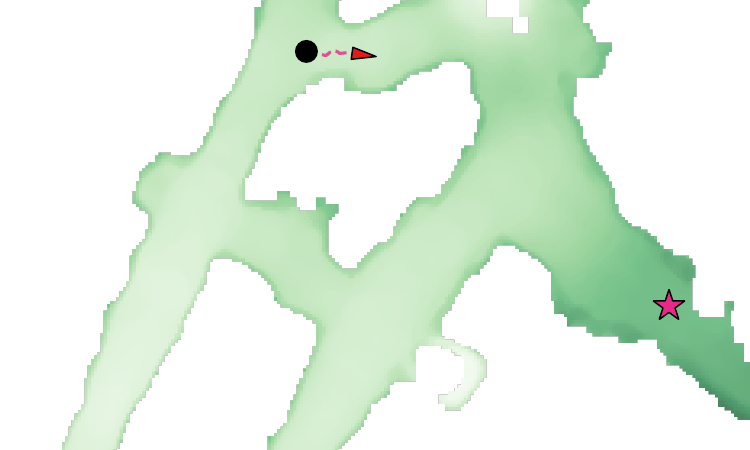}};
		\node[labelboxbl] at (b2c1.south west) {Elapsed: 7.38\,s\\Dist: 3.12\,m};

		\node[imagebox] (b1c2) at (\Cfive, \PAone)
		{\includegraphics[width=\W, height=\Hphoto]{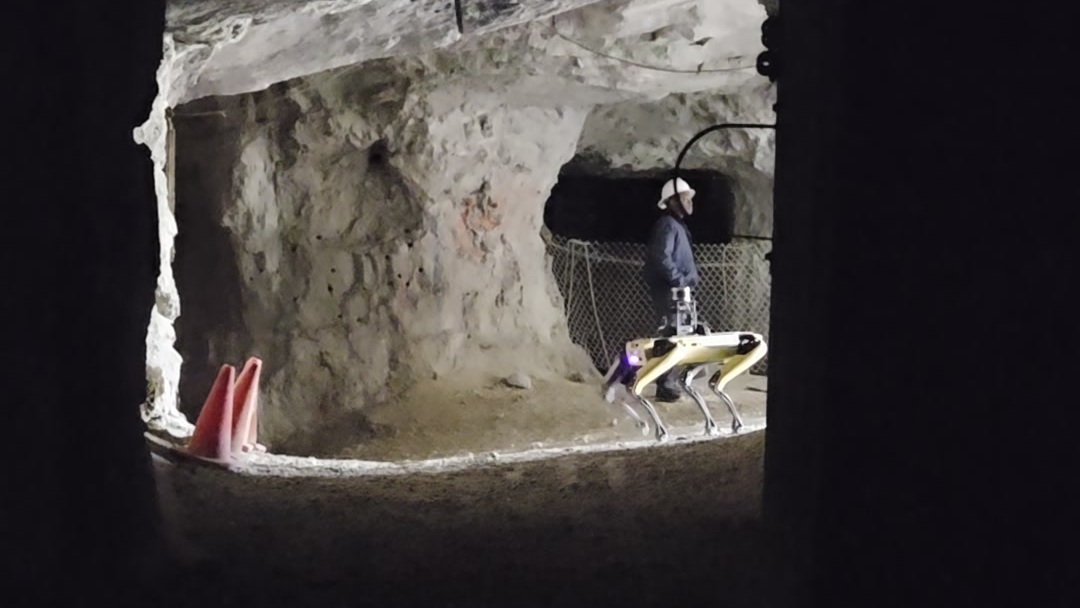}};
		\node[labelboxbl] at (b1c2.south west) {(e) $t_2$: Intersection};

		\node[imagebox] (b2c2) at (\Cfive, \PAtwo)
		{\includegraphics[width=\W, height=\Hmap]{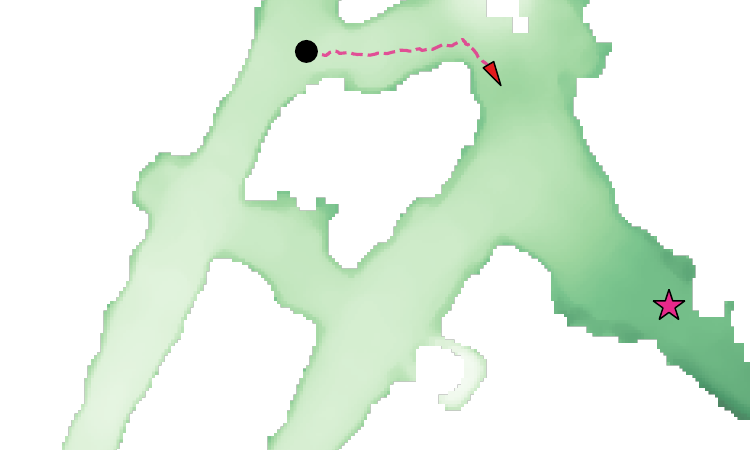}};
		\node[labelboxbl] at (b2c2.south west) {Elapsed: 25.38\,s\\Dist: 10.49\,m};

		\node[imagebox] (b1c3) at (\Csix, \PAone)
		{\includegraphics[width=\W, height=\Hphoto]{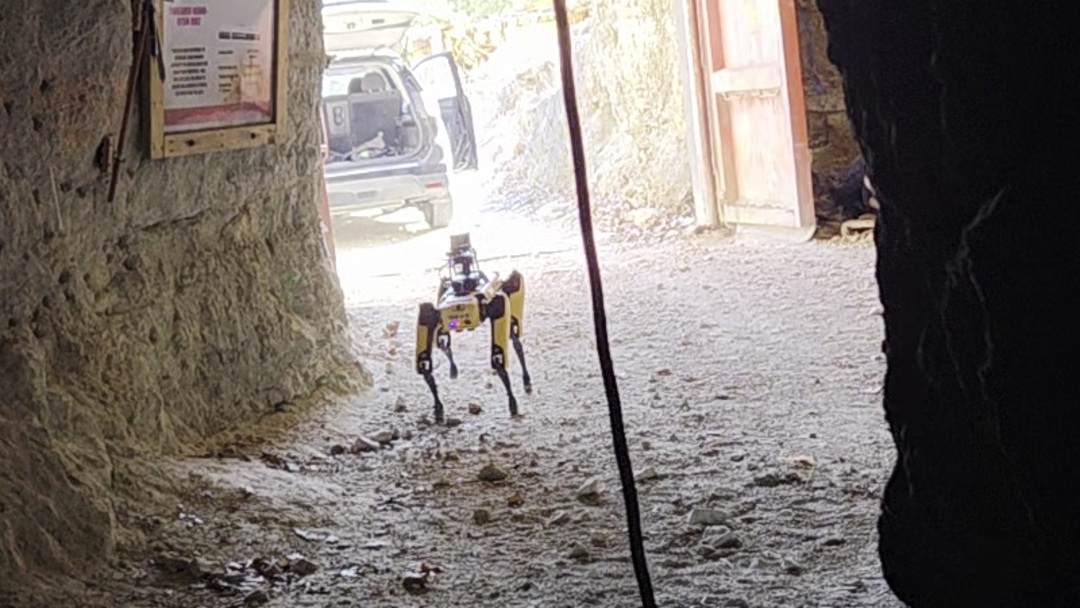}};
		\node[labelboxbl] at (b1c3.south west) {(f) $t_3$: Entrance Arrival};

		\node[imagebox] (b2c3) at (\Csix, \PAtwo)
		{\includegraphics[width=\W, height=\Hmap]{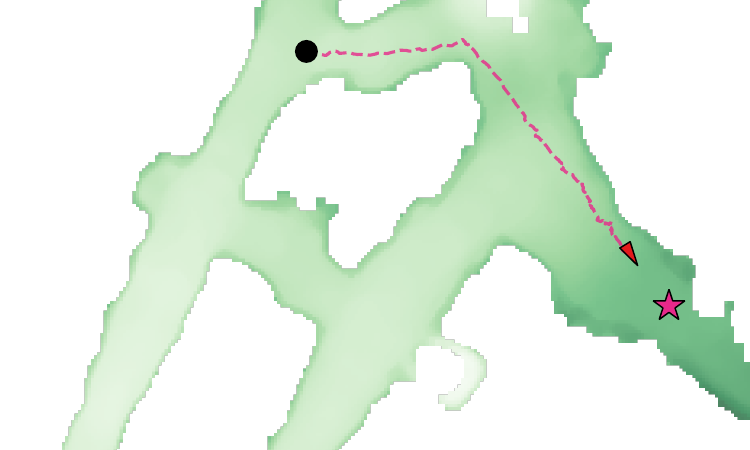}};
		\node[labelboxbl] at (b2c3.south west) {Elapsed: 69.38\,s\\Dist: 24.68\,m};

	\end{tikzpicture}
	\caption{Qualitative timeline and trajectory overview.
		\textbf{Left (a--c):} Goal~3
		(Entrance$\rightarrow$Deep)---the robot navigates through
		narrow, sloped, pitch-black passages.
		\textbf{Right (d--f):} Goal~4
		(Deep$\rightarrow$Entrance)---starting from a dark
		location, the robot returns to the entrance.}
	\label{fig:qualitative_timeline}
\end{figure*}
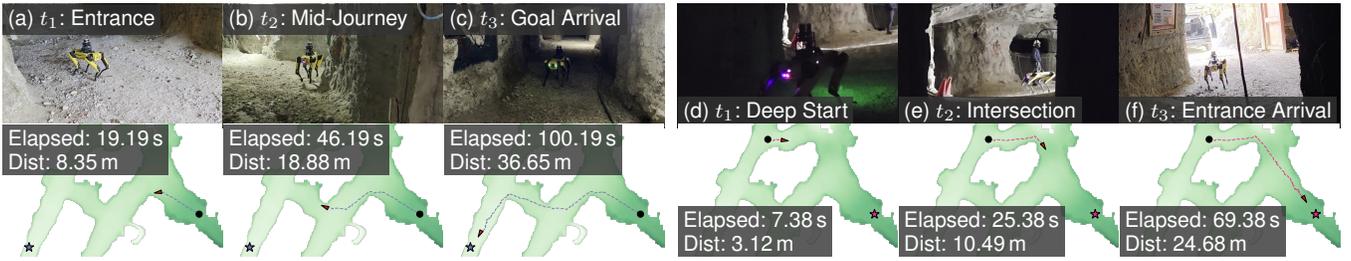

\begin{figure*}[t]
	\centering
	\includegraphics[width=\textwidth]{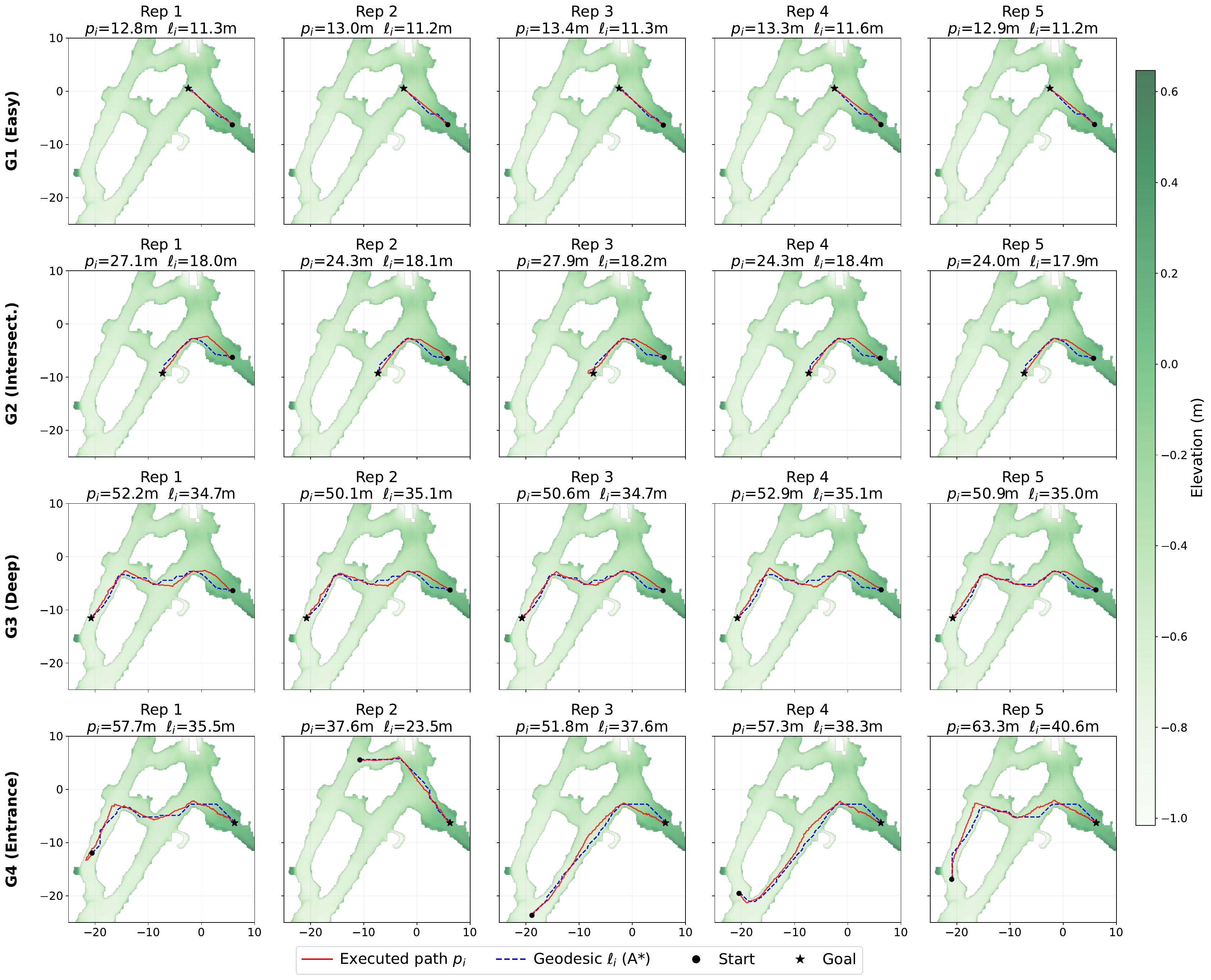}
	\caption{
		Executed paths (solid red) vs.\ geodesic shortest paths (dashed blue)
		on the elevation map for all 20 trials across the four target locations.
		$p_i$ is path length and $\ell_i$ is geodesic distance.
	}
	\label{fig:all_trials_grid}
\end{figure*}

We adopt the evaluation protocol of Anderson
et al.~\cite{anderson2018evaluation}, reporting success rate
(SR), success weighted by path length (SPL), path ratio
($p_i/\ell_i$), and completion time.

\begin{table}[t]
	\caption{Navigation performance (mean $\pm$ std over 5 repetitions per
		goal; 20 trials total). SPL per Anderson et
		al.~\cite{anderson2018evaluation}. ``All'' reports the mean of
		per-goal means.}
	\label{tab:results}
	\begin{center}
		\small
		\setlength{\tabcolsep}{2pt}
		\sisetup{separate-uncertainty}
		\begin{tabular}{@{} l c S[table-format=2.1(1.1)] S[table-format=1.2(1.2)] S[table-format=1.2(1.2)] S[table-format=3.0(2.0)] @{}}
			\toprule
			Goal        & Succ. & {Path (m)}   & {$p/\ell$}    & {SPL}         & {Time (s)} \\
			\midrule
			G1 (Easy)   & 5/5   & 13.1 \pm 0.2 & 1.16 \pm 0.02 & 0.87 \pm 0.01 & 27 \pm 1   \\
			G2 (Inter.) & 5/5   & 25.5 \pm 1.7 & 1.41 \pm 0.09 & 0.71 \pm 0.05 & 60 \pm 6   \\
			G3 (Deep)   & 5/5   & 51.3 \pm 1.0 & 1.47 \pm 0.03 & 0.68 \pm 0.01 & 127 \pm 7  \\
			G4 (Entr.)  & 5/5   & 53.5 \pm 8.8 & 1.53 \pm 0.09 & 0.66 \pm 0.04 & 131 \pm 20 \\
			\midrule
			All         & 20/20 & {35.9}       & {1.39}        & 0.73 \pm 0.09 & {86}       \\
			\bottomrule
		\end{tabular}
	\end{center}
\end{table}

\subsection{Navigation Performance}
\label{sec:nav_performance}

Table~\ref{tab:results} summarizes the results of all 20 trials.
The system achieved a 100\% success rate (20/20), with every trial
reaching within \SI{1.0}{m} of the target without human intervention.
The mean final distance to goal across all trials was \SI{0.25}{m}
(range: \SI{0.00} - \SI{0.66}{m}), well within the success threshold.

The overall SPL of $0.73 \pm 0.09$ (mean of per-goal means) reflects
a consistent trade-off between path optimality and the conservative
behaviors of the planner and controller. Goal~1, a short straight
traverse, achieved the highest SPL ($0.87$) with a path ratio of
only $1.16$, indicating near-optimal routing. As task difficulty
increases (longer distances, intersections, elevation changes) the
path ratio grows from $1.16$ (Goal~1) to $1.53$ (Goal~4), and SPL
decreases, respectively. This is expected: the FAR Planner routes
through visibility graph nodes that may not lie on the geodesic
shortest path, and the regulated pure pursuit controller introduces
additional path deviation through curvature-limited tracking.

Goal~4 (Entrance) exhibits the highest variance in path length
($\pm\SI{8.8}{m}$) and completion time ($\pm\SI{20}{s}$) because
each of the five repetitions starts from a different location in
the mine, resulting in geodesic distances ranging from
\SI{23.5}{m} to \SI{40.6}{m}. Despite this variability, SPL
remains tightly clustered ($0.66 \pm 0.04$), demonstrating that
the system generalizes reliably to arbitrary start-goal pairs
within the known map.

Completion times scale approximately linearly with path length
at an effective speed of \SI{0.38}{m/s}--\SI{0.48}{m/s}, below
the \SI{0.5}{m/s} maximum due to curvature and proximity
deceleration. The total autonomous traverse across all 20 trials
was \SI{717}{m} over approximately 29 minutes of navigation time.

\subsection{System Latency}
\label{sec:latency}

We measured the end-to-end perception-to-action latency by tracing each
velocity command back through the pipeline to the originating LiDAR scan
timestamp across all 20~trials. The pipeline is \emph{parallel}:  NDT
localization runs asynchronously and does not block the control path. The
critical path follows the terrain branch: VLP-16 driver
($\sim$\SI{84}{ms}) $\to$ FAST-LIO ($\sim$\SI{8}{ms}) $\to$ terrain analysis
($\sim$\SI{41}{ms}) $\to$ FAR Planner ($\sim$\SI{29}{ms}) $\to$ regulated pure
pursuit ($<$\SI{1}{ms}).

The resulting end-to-end latency from LiDAR scan to
velocity command has a median of \SI{176}{ms}
(mean \SI{195}{ms}), well within the
\SI{400}{ms} planning cycle, demonstrating consistent real-time
performance across all goal difficulties. NDT localization, which
provides global drift correction in parallel, has a median latency
of \SI{100}{ms} but occasionally takes up to \SI{1.1}{s} when
scan-matching converges slowly; because it runs asynchronously,
these outliers do not affect control responsiveness.

\subsection{Qualitative Analysis}
\label{sec:qualitative}

Fig.~\ref{fig:goals_combined} shows trajectories overlaid on the
mine elevation map for all four goals. Goals~1--3 share a common
start near the entrance, with the representative (median-SPL) run
shown for each. The trajectories closely follow the mine passages,
with deviations primarily occurring at intersections where the
planner selects a visibility graph route that differs slightly
from the geodesic path. Goal~4 demonstrates the system's ability
to plan from five distinct start positions throughout the mine,
all converging to the entrance.
Fig.~\ref{fig:all_trials_grid} expands this view to all
20 individual trials, confirming consistent path quality
across repetitions.

Fig.~\ref{fig:qualitative_timeline} presents a visual timeline of
two representative missions. The Goal~3 sequence (panels a--c)
shows the robot traversing from the well-lit entrance area into
progressively darker and narrower passages, reaching the deepest
target at \SI{36.7}{m} of travel in \SI{100}{s}. The Goal~4
sequence (panels d--f) illustrates the reverse: starting from a
completely dark zone deep in the mine, the robot navigates through
an intersection and returns to the entrance.
These sequences confirm that the LiDAR-only perception pipeline
operates identically regardless of ambient lighting, as the
3D point clouds provide full geometric information in both
the illuminated entrance area and the pitch-black interior.

\section{DISCUSSION}
\label{sec:discussion}

The 100\% success rate over 20 trials and \SI{717}{m} of
autonomous traversal demonstrates that reliable point-to-point
navigation in underground mines is achievable with classical,
interpretable components on a \SI{40}{W} edge computer. Several
design choices contribute to this reliability. Loading a
pre-computed visibility graph at startup eliminates the
exploration phase entirely, enabling immediate path planning to
any reachable goal. The two-stage localization pipeline (FAST-LIO2
for local consistency, NDT for global drift correction) maintains
accurate pose estimation throughout missions without the
computational overhead of full graph-based SLAM. The regulated
pure pursuit controller's curvature scaling and proximity
deceleration provide smooth velocity profiles suited to the
uneven terrain and tight turns of mine passages.

The system has several limitations. First, it requires a prior
map: a teleoperated mapping pass must be performed before
autonomous navigation, making the current system unsuitable for
unknown or dynamically changing environments. Second, the planner
does not handle dynamic obstacles; the mine was cleared of
personnel (except for the observer at the goal) during trials.
Third, while the PMF terrain segmentation and ceiling filter
parameters were tuned for this mine's geometry, transferring to
mines with significantly different passage dimensions or terrain
characteristics may require re-tuning. Fourth, the thermal camera
is currently used only for situational awareness and is not
integrated into the navigation pipeline.

To quantify the contribution of NDT localization, we analyzed
the \texttt{map}$\to$\texttt{odom} correction transform across
all 20 trials. NDT applies a median per-step XY correction of
\SI{1.1}{cm}, keeping accumulated drift bounded to
\SI{31}{cm}--\SI{103}{cm} depending on path length. Without
these corrections, FAST-LIO2 odometry alone would accumulate
sufficient drift on longer missions (Goals~3 and~4,
$>$\SI{35}{m}) to exceed the \SI{1.0}{m} success threshold,
particularly in the geometrically repetitive deep passages
where odometry drift is highest. Z-axis corrections are
larger (median \SI{3.1}{cm} per update) due to the low-ceiling
geometry providing weak vertical constraints for scan
matching, but this does not affect navigation since the
planner and controller operate in 2D.

\section{CONCLUSIONS}
\label{sec:conclusions}

We presented an autonomous navigation system for a quadruped
robot in underground mines that runs entirely on a \SI{40}{W}
edge computer without GPU acceleration, network connectivity,
or learned components. The system integrates LiDAR-inertial
odometry, NDT map-based localization, morphological terrain
segmentation, visibility-graph global planning, and regulated
pure pursuit into a real-time perception-to-action pipeline.
Field validation across 20 trials in an experimental underground
mine achieved a 100\% success rate and an SPL of
$0.73 \pm 0.09$, demonstrating that reliable autonomous
navigation in challenging subterranean environments does not
require high-power compute or learning-based methods.

Future work will: (1)~extend the
system to unknown environments through integrated exploration
and incremental map building, (2)~incorporate dynamic obstacle
detection and avoidance for operation alongside mine personnel and equipment, and
(3)~integrate the thermal camera for miner state and location tracking.

\section*{ACKNOWLEDGMENT}
\blackout{This research was funded by a grant from the Centers for Disease Control's
National Occupational Safety and Health (grant no. U60OH012350-01-00).}

\bibliographystyle{IEEEtran}
\bstctlcite{IEEEexample:BSTcontrol}
\bibliography{references}

\end{document}